\documentclass{article}
\usepackage{Odyssey2016,epsfig,amssymb,amsmath, enumitem, url, balance}

\setenumerate{itemsep=0pt, parsep=0pt}
\def\q#1{``\textsl{#1}''}

\def\fig#1{Figure~\ref{fig:#1}}
\def\eq#1{(\ref{eq:#1})}
\def\sec#1{Section~\ref{sec:#1}}

\def\x{\boldsymbol{x}}
\def\y{\boldsymbol{y}}
\def\s{\boldsymbol{s}}
\def\({\bigl(}
\def\){\bigr)}
\let\_=\underscore

\def\_{\ifmmode\let\next=\rmsub\else\let\next=\underscore\fi\next}

\def\ea{\textsl{et al.}}
\def\sn{\textit{Sprekend Nederland}}

\name{David A. van Leeuwen$^{1,2}$, Rosemary Orr$^2$}
\address{$^1$CLST/CLS, Radboud University Nijmegen, The Netherlands\\
  $^2$NovoLanguage, Nijmegen The Netherlands,\\
  $^3$University College Utrecht, The Netherlands\\
\texttt{d.vanleeuwen@let.ru.nl}, \texttt{r.orr@uu.nl}}

\title{The ``Sprekend Nederland'' project and its application to accent location}

\begin{document}

\maketitle

\abstract{This paper describes the data collection effort that is part of the project \sn\ (\textsl{The Netherlands Talking}), and discusses its potential use in Automatic Accent Location.  We define Automatic Accent Location as the task to describe the accent of a speaker in terms of the location of the speaker and its history.  We discuss possible ways of describing accent location, the consequence these have for the task of automatic accent location, and potential evaluation metrics. }

\section{Introduction}
\label{sec:introduction}

Automatic accent recognition from speech is traditionally treated as a classification task.  This matches quite well with our intuitive idea of accents, namely, the information in the speech signal that reveals the membership of a social group, e.g., fellow country members, people of the same social class, or ethnic background.  

One of the earlier attempts to cover several accent regions in the US for speech research was the collection of the TIMIT database~\cite{Lamel:1989}.  In this data collection, a total of 439 speakers were recruited from eight dialect regions~\cite{Zue:1990}.  This data was used by Hansen~\ea~\cite{Hansen:2004} for accent classification between seven of these dialect regions, where it was concluded that the TIMIT dialects were highly confusable with just two of the dialects, which attracted 60\,\% of all classifications.  A more recent study involving accent classification is carried out by Behravan~\ea~\cite{Behravan:2015}, where discrimination between eight foreign accents of Finnish was studied within an i-vector framework.  Bahari~\ea~\cite{Bahari:2013} also studied foreign accent classification, where the multilingual part of the Mixer database recordings from NIST Speaker Recognition Evaluations 2008 and 2010 was used to carry out accent recognition on five foreign accents of English.  In a comprehensive study Hanani~\ea~\cite{Hanani:2013} report on both automatic and human recognition of geographically close accents: 14 accents of the \emph{Accents of the British Isles} corpus, as well as two accents from different Ethnic groups of the \emph{Voices across Birmingham} corpus.  

Accent recognition was introduced in the NIST context of speech technology evaluations in 1996~\cite{NIST-lre-evalplan:1996}, where the distinction was made between the \emph{languages} General and Southern American English, mainland and Taiwanese Mandarin, and Caribbean and ``highland Spanish.''  In NIST Language Recognition Evaluations (LREs), the classification task is cast as a \emph{detection} task.  This is done in order to include a sense of calibration and to factor out the influence of the prior.  

In 2005 the interest in accent classes was revived, discriminating between American and Indian English, and mainland and Taiwanese Mandarin.  In 2007 two additional accent pairs were introduced: Caribbean and non-Caribbean Spanish, and Hindi and Urdu.  The latter contrast is often perceived as cultural or political, with many linguists considering the two languages to be the same. One of the LDC annotators noted that the similarity in recordings of Hindi and Urdu makes it very difficult to distinguish between the two.\footnote{As mentioned by George Doddington at the 2007 LRE workshop.}  Recently, in 2015, more focus to different accents has been given by defining \emph{language clusters}~\cite{nist-lre-evalplan:2015}. For the clusters Arabic, Chinese, and English, respectively, five, four and three regional accent variants are defined as separate classes to be recognized, as well as a cluster Iberian containing three regional accent variants of Spanish.  Since NIST LRE 2007 the language recognition task is not only evaluated in terms of detections costs, but also in terms of a logarithmic scoring rule assessing a probabilistic expression of the language classes~\cite{Brummer:2006a}. 

Human perception of accents is probably more a classification task than a detection task. From an evolutionary perspective, classification of accents might have advantages of quickly discriminating betweens friends and foes, and of assessing potential threats and opportunities in dealing with people from outside the local community.  When we hear someone speaking whom we do not know, the first intuitive question  which arises is: \emph{Where is this person from?}  In this question ``where'' is not limited to geographical location, but includes social communities.  In human perception, the classes and priors vary from listener to listener, and will be influenced by the mobility of the listener and exposure to speech from other social groups. 

In this paper we would like to generalize the accent classification task to one that is closer to the intuitive human perception question ``where does this person come from?''.  We will do this along the lines of the information that is collected in the Dutch project \sn, which might be translated freely as ``The Netherlands Talking.''  \sn\ is a project whose goal is to make an inventory of the accents spoken in the Netherlands, and research the attitudes of people towards others with a different accent.  Data is collected by means of an application (app) on a smartphone, which implements speech acquisition and playback, and metadata and attitude entry.  The metadata that is important for this paper is information about where people lived during their life, as well as what accent others perceive them as having.  Additionally, speech recordings are accompanied by the device location at the time of the recording, which may reveal information about the mobility of the participants.  The data is being collected over a period of a year.  Two months after the start we have about 5000 participants.  The data is very heterogenous, with varying amounts of audio and metadata per participant.  This is a result of the inclusive design and voluntary nature of participation.  

Accent data collection from multiple regions can be problematic, as was shown by Bock and Shamir~\cite{Bock:2015} who used the eight regional accents in the Voxforge crowdsourcing effort.  The authors were able to show perfect accent classification by just using a second of silence from each Voxforge recording.  Clearly this is not an effect of accent, but of regional difference in recording quality or technical parameters.   
 
This paper is organized as follows.  First we will describe the efforts and the data characteristics of the \sn\ project, and then formally introduce the task of accent location that can be researched with this kind of data.  We will conclude by discussing potential evaluation metrics for this task. 

\section{The Sprekend Nederland project}

\sn\ is a project initiated by the Dutch broadcast organisation NTR, and is partly inspired by the UCU Accents project~\cite{Orr:2011} that studies accent convergence in a community of undergraduate students in an international College.  \sn, however, has its focus on accent variability and attitude towards spoken accents.  The NTR is supported by a group of researchers from various disciplines, who see the opportunities that are presented for their own research by this data collection effort.  The intention is to make the data available to the research community at the end of the project. 

\subsection{Data acquisition}
\label{sec:data-acquisition}

There are three primary kinds of data collected in the project: audio material (spoken utterances), attitude data (judgements about other participants' speech) and metadata (information about the participants themselves).  All three forms of data are collected using an \emph{app} that runs on a smartphone or tablet.  The app is developed by a third party for the NTR, as a native application to the iOS and Android platforms.  

Audio is recorded through the device's current microphone(s), which usually will be the built-in microphone but may also be a head set or remote microphone.  One- or two-channel recordings are made at 44.1 kHz and compressed using a conservative AAC compression at approximately 110\,kb/s.  This format is somewhat alien to the speech community, but it was chosen by the app developer to enhance cross-device interoperability.  Informal listening suggest that the recording quality of current day smartphones and tablets is actually very high.  The start of the recording is initiated by the participant, who also controls the end of the recording in most cases.  After a recording, there is an opportunity to play back the recording just made, and if the participants are not satisfied with the first attempt(s) they can re-record an utterance.  

Metadata and attitude data are acquired by prompting a question, and recording the action in one of four response categories:
\begin{enumerate}
\item A discrete slider, with labels at the minimum (left) and maximum (right), with optionally a current value above the slider position.
\item ``Yes'' or ``No'' buttons
\item A list of items that each can be selected individually
\item A location on a map, with free pan and zoom
\end{enumerate}

For \emph{metadata}, the question is about the participant, and there are currently 53 metadata questions.  For this paper, potentially relevant questions are: \q{Have you always lived at the same location?}, \q{How long ago did you move to your current location?}, \q{Where have you lived the longest within the Netherlands?}, \q{Where do you come from?}, \q{Where did you go to primary school?}, \q{Where did you go to secondary school?}, \q{What is your native language?}, \q{Do you speak a Dutch dialect?}, \q{Where would you locate your accent on a map?}, as well as demographic questions about sex, age, etc.  

For \emph{attitude data} the participant first listens to someone else recorded earlier.  Then, attitude questions like \q{Would you like to have this person as your neighbours?} follow.  For this paper, relevant questions in this category of 38 questions are: \q{Where does this speaker come from?}, \q{Is this a native speaker of Dutch?}, \q{How far away from you do you think the speaker lives?}, \q{Can you locate the speaker on a map?} and perhaps \q{How old do you think this person is?}.  

\subsection{Recruitment of participants}
\label{sec:recr-part}

Participants are recruited through traditional and social media.  Potential participants are encouraged to install the app on their telephone.  After installing, the app displays several pages in which the overall intention and operation is conveyed. Further use of the app requires registration.  The purpose of registration is to ensure that each participant's data is associated with that participant only.  To this end, a participant provides an e-mail address.  However, in the implementation, this e-mail address is separated from other data in such a way that the researchers working with the data do not have access to this information---everything is stored under a numeric participant ID.  A third party has access to the email-adresses to handle password resets and requests to be removed from the database.  With registration participants agree to terms and conditions, and approve the use of data entered in the app for use in the project and for scientific research.  For participants under 18, parents or guardians are required to agree to the terms and conditions.  

\begin{figure}
  \centering
  \includegraphics[width=\hsize]{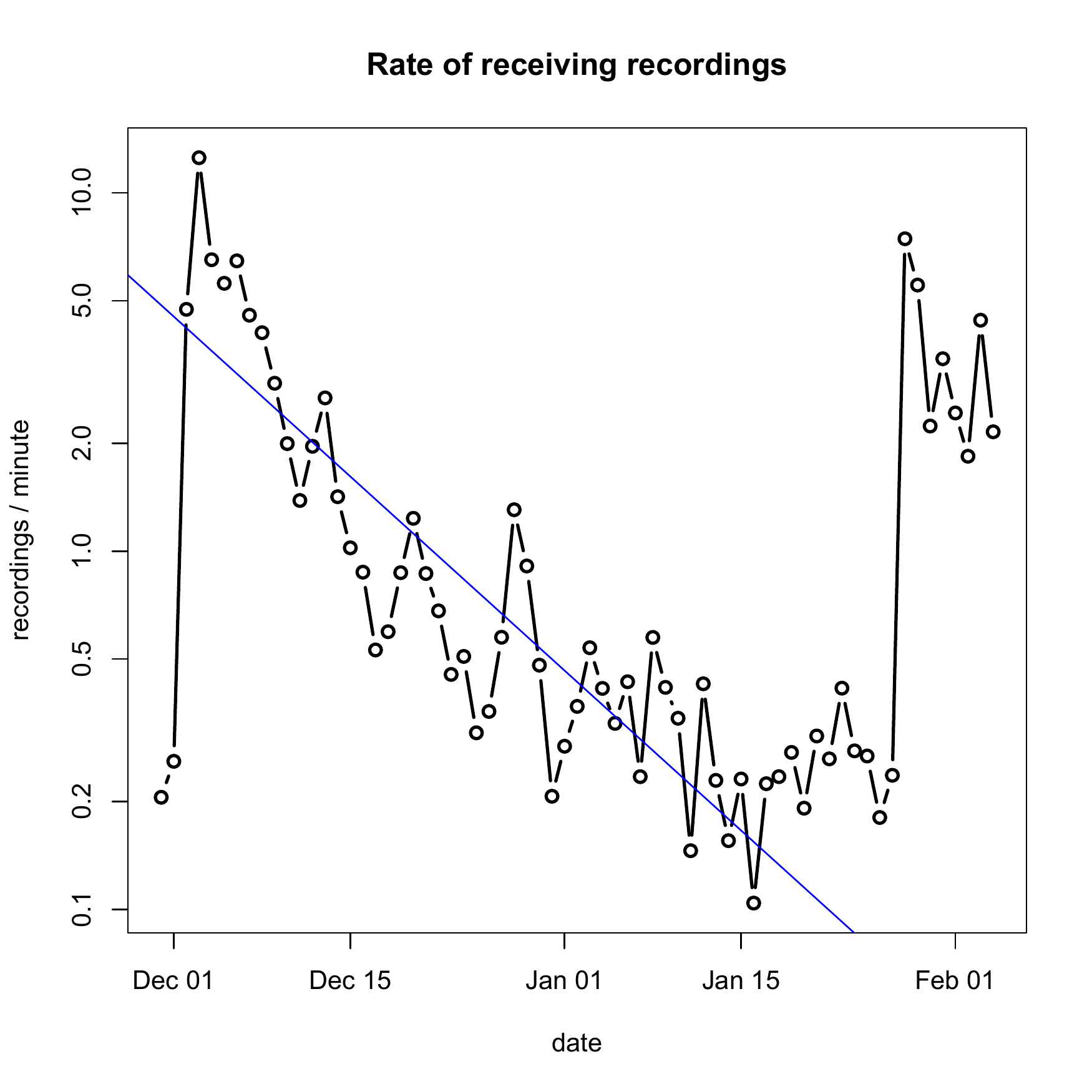}
  \caption{Rate at which speech recordings are made right after the launch.   Data points represent averages over calendar days, the fitted line represents exponential decay from 2 Dec 2015 (the launch) until 18 Jan 2016.}
  \label{fig:recording-rate}
\end{figure}

Traditional media events, such as the mentioning of the app on public radio and television broadcasts, appears to be a dominant factor in the recruitment and activity of participants.  This may be appreciated from \fig{recording-rate}, which shows the daily rate at which recordings were made after an initial launching which was accompanied by some exposure on TV and radio. A second boost in recording rate occurred after a documentary on national television about the project at 28 Jan 2016.  

\subsection{Spoken content and sessions}
\label{sec:spoken-content}

Several kinds of stimulus material are used in order to elicit speech.  We drew our material from six lists of items.  These lists were composed by several researchers from different research institutes across the Netherlands. List~1 consists of 10 sentences and 44 words, which, together, cover all known regional phonemic variability in the country.  List~2 consists of 122 loan words, for which it is most likely that they display a speaker's natural accent.  List~3 consists of 100M sentences drawn from the COW corpus (``COrpus from the Web'')~\cite{COW:2015} for maximum lexical variability, but was toned down to about 2000 sentences for operational reasons.  List~4 consists of 66 hand drawn pictures, used to elicit regional varying lexical items.  List~5 consists of 204 words which cover all possible consonant-vowel pairs that are used in Dutch.  Finally, list~6 consists of nine description tasks used to elicit spontaneous speech. 

Stimulus material consisting of sentences is presented as a reading task.  Material consisting of words is presented as a paced reading task, where at fixed time intervals, the next word is shown in groups of five words.  The pictures are shown a short while after which the item must be spoken within a fixed time frame.  

Each list has its own target regarding completeness: for List~1, the goal is that every participant completes all sentences and words.  For the other lists, material is randomly sampled for each participant.  Because of the large number of items to be recorded, all stimulus material is categorized by topic, where there are six possible topics.  These topics quite naturally get the interpretation of a \emph{session}, where recordings, attitude questions and metadata questions are mixed in order.  Because it takes a fair amount of time to complete a topic (about 20 min), we expect participants to pace the completion of all topics over a longer period of time.  In this way, we hope to cover extended periods of time and different sessions for the same speaker.  

Where in speaker recognition research, it is important to have recordings of the same speaker through different channels and recording devices, such variability is expected to be small in this collection.

\section{Some early statistics of participation}
\label{sec:some-early-stat}

In this section we will present some early results in the data collection, as inspiration for potential speech technological tasks related to accent recognition that we can use on the data.  

The data from \fig{recording-rate} shows the participation after the initial launch.  The exponential fit is made from the launch media event until 18 Jan 2016, where a video about the project was posted on social media, and was viewed about 3k times.  Assuming an exponential drop at $d=7.1\,\%$ per day, as shown by the fitted line, we can make a prediction for the total number of recordings associated with this event as $NR_0/(1-d)$, where $R_0$ is the initial rate on the first day after the event, and $N=24\times 60$ is the number of minutes in a day.  
To optimize the amount of data acquired, it is important to keep $d$ low via an appealing app design and user experience, and effective use of social media.  

The rate at which participants register and the rate of answers given follows a similar pattern, with decay constants of 8.3\,\% and 6.6\,\% per day. 

\begin{figure}
  \centering
  \includegraphics[width=\hsize]{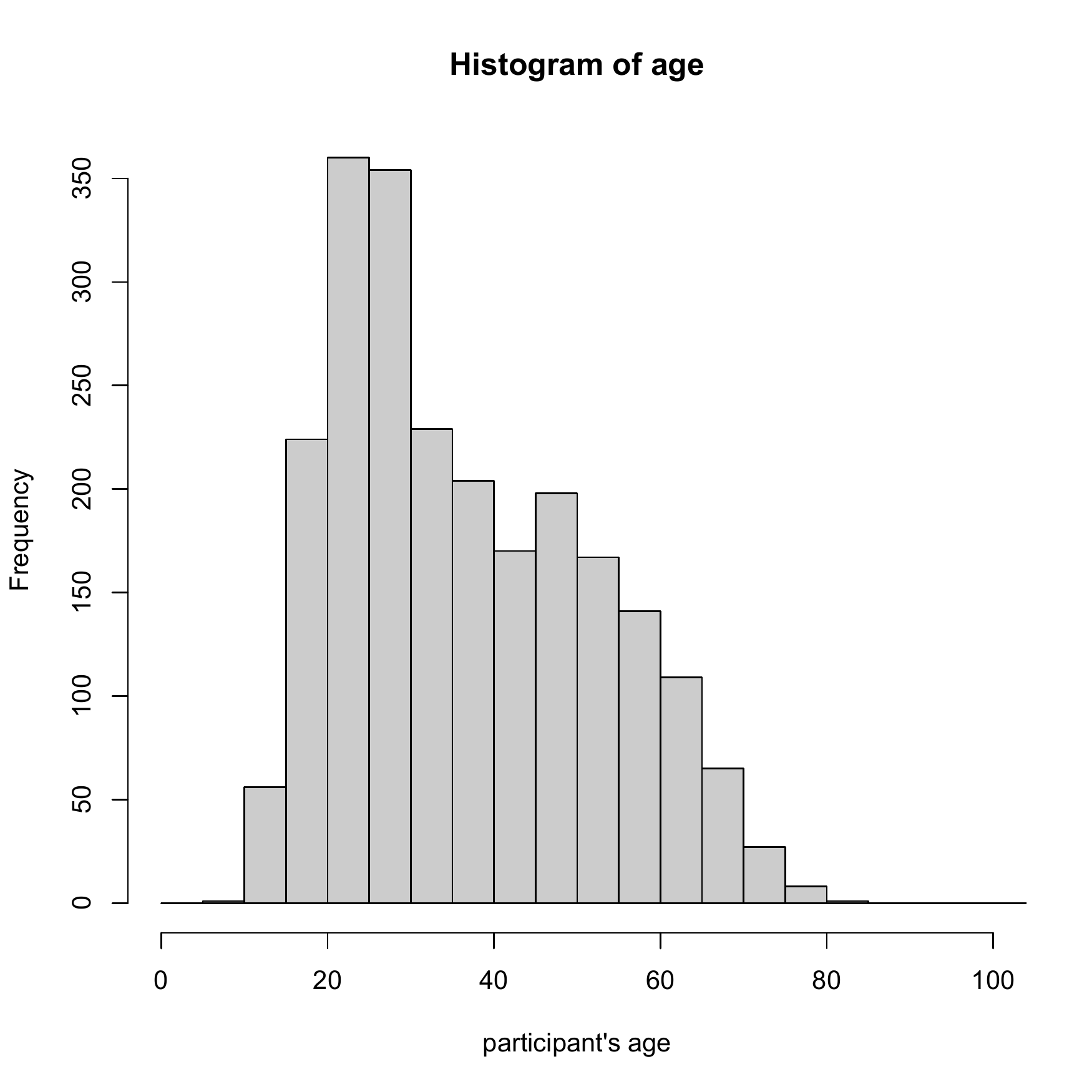}
  \caption{Age distribution of registered participants}
  \label{fig:age-dist}
\end{figure}

The age distribution in \fig{age-dist} shows that young adults have the highest representation in the data.  We only have the birth year of 31\,\% of the participants.  There are two likely reasons for this.  Firstly, any of the questions can be skipped if the participant wants, and secondly, all 53 metadata questions are distributed over all six topics, and participants may have not reached the question yet, or even stopped using the app.  Participants can of course lie about their age, but distribution does not show accumulation at extreme values which might be expected in the case of untruthful answers.  Also, to the question ``Did you answer all questions truthfully'' (ignoring the philosophical issues of this question for now) only 0.5\,\% answered negatively.  The gender balance is 57/43\,\% female/male, with only 0.3\,\% of the participants indicating ``other.''  

\begin{figure}
  \centering
  \includegraphics[width=\hsize,viewport=60 100 350 420,clip]{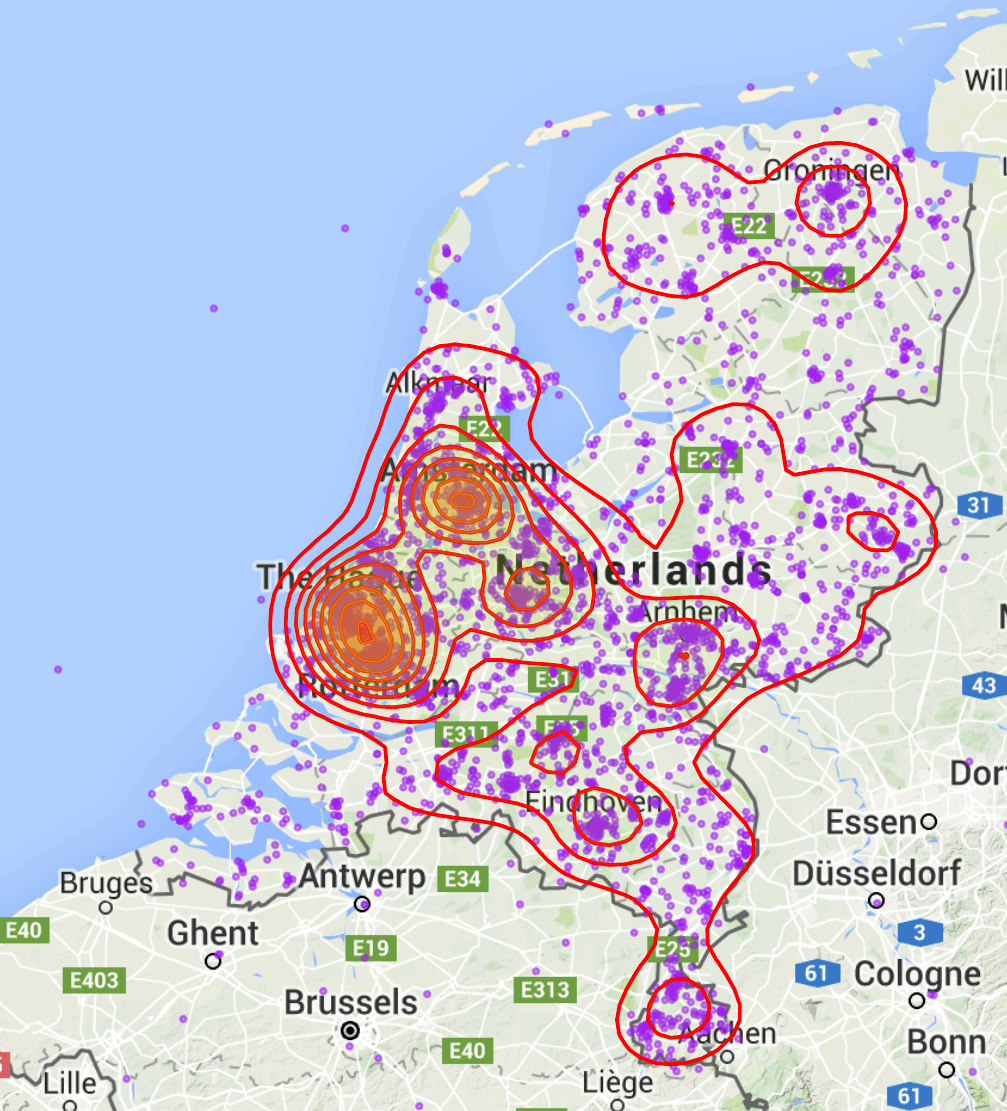}
  \caption{Self-reported answer to the question \q{Where do you come from?}, as data points (purple) and density contours (red).  Graph produced using~\texttt{ggmap}~\cite{Kahle:2013}.}
  \label{fig:q04}
\end{figure}

About 50\,\% of the participants gave an answer to the question \q{Where do you come from?}.  The answers are given as a location on a map, where the participant is free to use the pan and zoom capabilities of the smartphone's map interface.  These locations are summarized in \fig{q04}, both as raw data and density contours.  Assuming that the density of population is reflected in the density of the participants, this supports the idea that participants mostly answer honestly.  It is striking that hardly any participants appear to be from Belgium, where over half the population speaks Dutch.  This is probably a result of the cultural divide between the countries, which have separate radio and TV channels and newspapers.  The project name, of course, does not immediately suggest that speakers from Belgium are invited to participate.

\section{Accent location}
\label{sec:accent-location}

Individual accents are largely formed by interaction with one's social group.  When these groups are primarily distinguished by social class, we might call the accent a \emph{sociolect}, and when the groups are formed by cultural background, e.g., an immigrant population from a common origin, we might call this an \emph{ethnolect}.  Although the set-up of the project is general enough that evidence of both sources of accents can be found, we will be concerned here only with accent as a result of location.  

We will now explore a formal definition of the information that was shown in \fig{q04}, the answer to the question \q{Where do you come from?}, which we will define as the \emph{origin location}~$L$.  For a person who has lived her entire life at the same location $\x$ (a vector, e.g., longitude/latitude coordinates), we might define the origin location by
\begin{equation}
  \label{eq:1}
  L_1:\quad \x .
\end{equation}
However, this person will have interacted with people in the neighbourhood, and her accent will be formed by these interactions.  So it might be better to define where she comes from as a probability distribution, that peaks around the location she lives
\begin{equation}
L_2:\quad p(\x).  
\end{equation}
For someone who has moved about in her life, we should include the location history in the distribution, as in 
\begin{equation}
  \label{eq:pdf-xt}
  L_3:\quad p\(\x(t)\).
\end{equation}
Here $t$ measure time along her life span.  At first sight, this may appear to be an over-cautious definition of origin location.  However, \eq{pdf-xt} is just a general definition, and an actual case might be simplified in description.  For instance, the dependence on $t$ may be specified as time intervals over periods where someone lived in the same place, and the distribution component over $\x$ may be chosen as a Gaussian, or even a $\delta$-distribution.  

In every day parlance, we can talk of accents along the lines of definition $L_3$.  For instance, somebody from Germany who has lived many years in New York may be recognized as such from her accent.  In forensic speaker analysis, an examiner may express traits of a perpetrator's accent as having traces from different regional accents, which also indicates a history like~$L_3$.  Note that the probability distribution in \eq{pdf-xt} can be generalized to include social groups or other accent-influencing factors. 

In order to arrive at a more point-like specification of the origin location, we might integrate over time
\begin{equation}
  \label{eq:3}
  L_4:\quad \int_t w(t) p\(\x(t) = \x\) \, dt, 
\end{equation}
where $w(t)$ is a weighting function that incorporates the susceptibility to picking up regional accents at a certain age.  Further marginalizing over position, we arrive at 
\begin{equation}
  \label{eq:4}
  L_5:\quad \int_t \int_{\x} w(t) \, v(\x) \, p\(\x(t)=\x\) \, \x \, d\x\, dt,
\end{equation}
where $v(\x)$ expresses a relative strength that a local accent has on people exposed to that accent.  Definition $L_5$ is point-like, similar to $L_1$, but represents some averaged location.  Such a description may literally ``miss the point'' if $L_4$ has multiple modes located far apart.  

\subsection{Accent location task}
\label{sec:accent-location-task}

In the accent location task, we want to know what the origin location~$L$ of a speaker is, given one or more spoken utterances~$\s$.  In its most general form, using definition~$L_3$, the task can be expressed as computing the posterior distribution
\begin{equation}
  \label{eq:post-xt}
  p\(\x(t)\mid \s\).
\end{equation}
Admittedly, this is quite a phenomenal task.  However, if we generalize accent location to language, then just recognizing the language~$l$ from $\s$ already narrows the posterior down to something like the prior
\begin{equation}
  \label{eq:6}
  \pi_l\(\x(t)\),
\end{equation}
which can be found by studying demographics statistics in the country where the language is spoken.  An example of what $p\(\x(\hbox{anno 2016})\)$ could look like if the recognized language is Dutch is shown in \fig{q04}.  

A similar reasoning holds for the accent location task: in effect it is the task of narrowing down the prior, given the evidence~$\s$.  

In most cases we would probably like to integrate over time, so that the task becomes slightly less daunting and we are effectively using an origin location definition like $L_4$ or $L_2$.  We can express the posterior now as
\begin{equation}
  \label{eq:post-x}
  p(\x\mid\s) = \frac{p(\s\mid\x)\,\pi(\x)}{\pi(\s)},
\end{equation}
which shows the importance of the prior~$\pi(\x)$---the population density---again.  

\subsection{Point estimates and regression}
\label{sec:point-estim-regr}

In most recognition tasks we tend not to give an answer in terms of a probability distribution, but rather as single value, a class or a scalar or vector.  One way of producing a vector is by integrating over position
\begin{align}
  \x\_{hyp}(\s) &= \int_{\x} p(\x\mid\s)\,\x\,d\x,\label{eq:x-hyp}\\
  \x\_{ref} &= \int_{\x} p(\x)\,\x\,d\x.\label{eq:x-ref}
\end{align}
Here, we have made point estimates of both the prediction (or hypothesis) and the reference distribution.  In this way, the accent location has turned into a regression problem.  Instead of the mean, we could have chosen the maximum of the posterior, i.e., the location of the highest mode.  

The advantage of turning accent location into a regression task is that it is easy to choose an evaluation metric: something based on the distance between $\x\_{hyp}$ and $\x\_{ref}$ would make sense.  A disadvantage, however, is that for someone who has lived in two distant locations, e.g., the East and West coast of the US, the mean location (at the center of the US) does not seem to be representative of this person's origin location.  

\subsection{Regional aggregates and classification}
\label{sec:regi-aggr-class}

\begin{figure}
  \centering
  \includegraphics[width=\hsize]{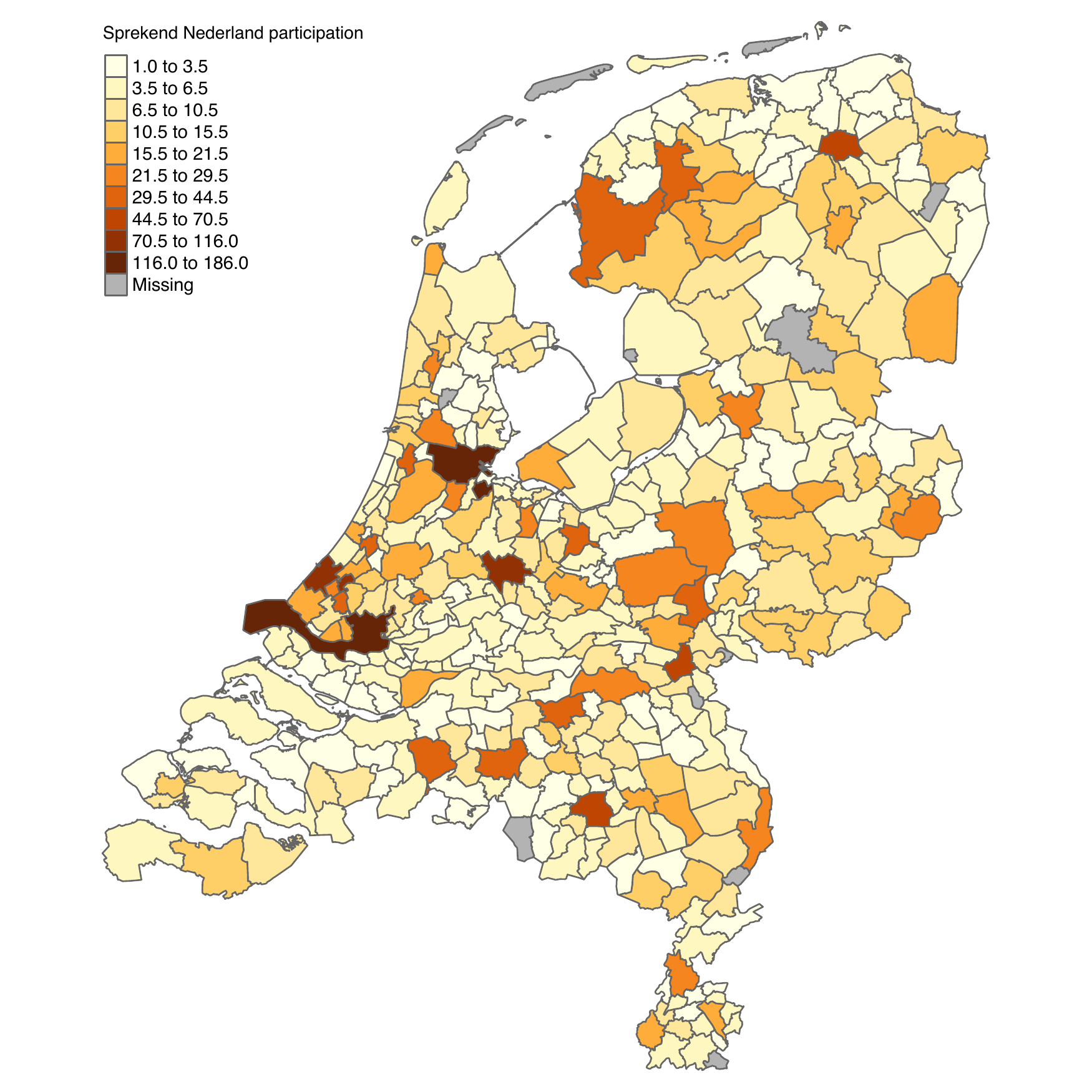}
  \caption{Number of participants per municipality in the Netherlands.  The darkest regions correspond to the cities Amsterdam and Rotterdam, respectively.  Graph produced using \texttt{tmap}~\cite{Tennekes:2015}.}
  \label{fig:counts-muni}
\end{figure}

A middle ground between a full spatial distribution and a point estimate is an integration over a geographical region.  Examples of such regions are city districts, municipalities, provinces, and countries. These are cultural-political entities, but for our analysis, any tessellation of the location space will do.  If $R_i$ is a set of geographical regions covering the location space, the hypothesis and reference can be turned into discrete distributions
\begin{align}
  p\_{hyp}(i\mid\s) &= \int_{\x\in R_i} p(\x\mid\s)\,d\x\label{eq:p-hyp}\\
  p\_{ref}(i) &= \int_{\x\in R_i} p(\x)\,d\x\label{eq:p-ref}
\end{align}
An example of the $\delta$-distributions formed by the purple dots in \fig{q04} integrated over a tessellation formed by the Dutch municipalities is shown in \fig{counts-muni}.  The discrete probabilities have been scaled by the total number of participants, so that in fact counts per municipality are shown.  This graph shows participants to \sn, but of course if one wants to set a region prior $\pi(i)$ it is better to use population statistics.  

With a probability distribution over regions, it is easy to do accent \emph{classification}, by simply choosing the region with maximum posterior probability:
\begin{align}
  \label{eq:9}
  c\_{hyp} &= \arg\,\max_i \, p\_{hyp}(i\mid\s),\\
  c\_{ref} &= \arg\,\max_i \, p\_{ref}(i).
\end{align}
This formulation is what is usually used in accent recognition research.  

\section{Evaluation metrics}
\label{sec:evaluation-metrics}

With our general task \eq{post-xt} we have not only set a challenge to the engineer, who has to devise methods to produce the required distribution. With the corresponding definition of origin location~\eq{pdf-xt}, we have also made it hard for an evaluator to find out how well an accent location system performs, and for the user, who has to interpret this.  In this section we will consider some evaluation metrics that may be appropriate to the various tasks defined in \sec{accent-location}.  

We want to concentrate on properties of an evaluation metric that are important for developing an accent location system, i.e., a better value should somehow represent a more useful system in general.  Properties to consider are
\begin{enumerate}
\item Probabilistic interpretation.  A probabilistic statement about a location should not be an under- or overestimation when evaluated over a collection of trials. 
\item Sense of distance.  A hypothesis location closer to a reference location should be evaluated as better. 
\end{enumerate}

Bahari and Van hamme~\cite{Bahari:2014a} have a comprehensive overview of evaluation metrics that consider the \emph{order} of classes, e.g., in age groups, as well as the probability of the classes.  They also introduce a metric called Normalized Ordinal Distance, and show by examples how this has better properties than existing metrics.  A completely different approach is made by Br\"ummer and Du Preez~\cite{Brummer:2006} who focus on the probabilistic specification of classes and calibration.  An important concept in this work is a \emph{proper scoring rule}, a function whose expected value is optimum when the predicted probability distribution is equal to the actual probability distribution.  The NIST primary evaluation metrics $C\_{det}$ and $C\_{primary}$ in speaker recognition, as well as $C\_{avg}$ in language recognition, are examples of proper scoring rules.  These metrics probe the calibration abilities of the predictor in one or a few points in the space of decision cost functions.  By contrast, metrics based on logarithmic scoring rules such as $C\_{llr}$~\cite{Brummer:2006, Appindep-eval:2007} integrate over the entire space of decision cost functions\footnote{This happens with a particular weighting, which might not correspond to the prior interest of the user.}.  A generalization of $C\_{llr}$ to more than one class was made in~\cite{Brummer:2006a} for language recognition, and the derivation of the logarithmic scoring rule for multiclass systems can be found in~\cite{Brummer-PhD:2010}.  For predictions of continuous probability density functions, a logarithmic scoring rule \emph{negative log predictive density} was defined as primary evaluation metric in the \emph{Evaluating Predictive Uncertainty Challenge}~\cite{Quionero:2006} of the PASCAL project.  In the followup of the challenge, Kohonen and Suomela~\cite{Kohonen:2006} showed some serious flaws in this metric, and proposed a metric that is sensitive to distance, the \emph{continuous ranked probability score}.  These metrics were studied in assessing predicted probability densities in age recognition~\cite{age-calibration:2012} and were further analyzed in terms of minimum age detection. 

\subsection{Local metrics}

We start with considering a metric for the case of regional aggregates and classification of~\sec{regi-aggr-class}.  When both reference~\eq{p-ref} and hypothesis~\eq{p-hyp} are a probability distribution over classes, a candidate performance metric is the cross entropy between reference and hypothesis distributions, averaged over the trial set~$\cal T$
\begin{equation}
  \label{eq:xent}
  H\_{region}(\hbox{ref, hyp}) = - \frac1{|\cal T|} \sum_{t\in\cal T} \sum_i p^t\_{ref}(i) \log p^t\_{hyp}(i).
\end{equation}
This metric is called \emph{local} because no benefit is gained by having posterior weight in regions neighbouring the regions where the true probability density is.  The cross entropy is an error metric (low values are better), is non-negative, and only zero if $p\_{ref}$ is 1 for exactly one region, and $p\_{hyp}$ as well, for the correct region.  However, if $p\_{hyp}$ vanishes for classes with finite $p\_{ref}$, the error can be unbounded.  

The prior distribution is of influence to the cross entropy.  A reference value could therefore be formed by inserting the prior $\pi(i)$ in place of $p\_{hyp}^t(i)$ in \eq{xent}.  If the set of trials are drawn from the same prior distribution, this reference value reduces to the entropy of the prior.  This entropy $H(\pi)$ for the data shown in \fig{counts-muni} is 5.40, for real population statistics it is 5.48.  

We can also aim at removing the influence of the prior.  This would lead to something similar to multiclass $C\_{llr}$~\cite{Brummer:2006a}, except that our reference is formed by a distribution over accent regions, rather than being limited to a single region.  It is debatable whether the prior should be factored out in accent location, as we are used to in speaker and language recognition.  We might take the stand that knowing where people live is part of the task of figuring out where someone comes from, just like knowing the prior over word sequences is considered part of the task in speech recognition. 

\subsection{Distance-sensitive metrics}

When the accent location task is formulated as a regression problem~\eq{x-ref}, we can easily build in distance-sensitivity in the evaluation metric.  We can simply define a distance-based error function
\begin{equation}
  \label{eq:E-regression}
  E\_{regression}(\hbox{ref, hyp}) = \frac1{|\cal T|} \sum_{t\in\cal T} D(\x\_{ref}^t, \x\_{hyp}^t),
\end{equation}
where $D(\x,\y)$ is some distance function dependent on $\x$ and $\y$.  The Euclidean distance is a viable candidate, but we can think of more advanced functions as well.  For instance, we may want $D$ to be saturated when the Euclidian distance becomes too large, or we might want to let $D$ depend on the population density between $\x$ and $\y$.  

For an accent location task of~\eq{post-x} with continuous probability density functions, we would like to assess both probability calibration and distance.  We know that a simple log scoring rule like the negative log predictive density can be misused~\cite{Kohonen:2006} and is further not distance-sensitive.  The continuous ranked probability score~\cite{Kohonen:2006}, and also normalized ordinal distance~\cite{Bahari:2014a}, work by comparing cumulative distribution functions between reference and hypothesis.  For a scalar predictor with natural order, like age, such cumulative distribution can be made, but for a two-dimensional predictor, location, it is not immediately clear how we can generalize this. 

One metric that is worth investigating, is the integral over both probability density functions, weighted by a distance function.  For a single trial this would be along the lines of
\begin{equation}
  \label{eq:E-dist}
  E\_{dist} = \int_{\x} \int_{\y} p(\x \mid \s) \, p(\y) \, D(\x, \y) \,d\x\, d\y.
\end{equation}
Such a metric is not ideal.  Even if $p(\x\mid\s) = p(\x)$, i.e., the recognizer predicts the accent distribution exactly, the error is not zero, and will depend on the reference $p(\x)$.  It will take more research to study the properties of~\eq{E-dist} and similar distance-aware metrics for comparing probability distributions.  

\section{Conclusions}
\label{sec:conclusions}

In this paper we have introduced a new task in speech technology, \emph{Accent Location}.  This is inspired by the descriptive statistical analysis of the on-going data collection project \sn, in which already over 5000 people are participating.  The project may be one of the first of its kind collecting speech, attitude and metadata on a large scale using modern mobile telephones and utilizing both traditional and social media for recruiting participants.  We believe that this data collection makes it possible to research accent location as a speech technological task.   Accent location in its most fundamental form will be very challenging, depending on what exactly is required in the task and what the geographical accent variation and distribution of speakers is.  We have further explored a number of evaluation metrics that may be suitable for specific accent location tasks. 

\balance

\bibliographystyle{IEEEbib}
\bibliography{david-bibdesk}

\balance

\end{document}